\documentclass[11pt,a4paper]{article}

\usepackage[T1]{fontenc}
\usepackage[utf8]{inputenc}
\usepackage{natbib}
\usepackage{booktabs}
\usepackage[english]{babel}

\usepackage{tabularx}
\usepackage{siunitx}
\usepackage{caption}

\usepackage{amsmath, amssymb}
\usepackage{multirow}
\usepackage{placeins}

\usepackage{graphicx}

\usepackage[hidelinks]{hyperref}

\title{Variational Distributional Neuron}

\author{
  Yves Ruffenach \\
  Conservatoire National des Arts et M\'etiers \\
  \texttt{yves@ruffenach.net} \\
  \href{https://orcid.org/0009-0009-4737-0555}{ORCID: 0009-0009-4737-0555}
}

\date{}

\begin{document}

\maketitle

\begin{abstract}
We propose a proof of concept for a \emph{variational distributional neuron}: a compute unit formulated as a VAE brick, explicitly carrying a prior, an amortized posterior, and a local ELBO. The unit is no longer a deterministic scalar but a distribution: computing is no longer about propagating values, but about contracting a continuous space of possibilities under constraints. Each neuron parameterizes a posterior, propagates a reparameterized sample, and is regularized by the KL term of a local ELBO --- hence, the activation is distributional. This ``contraction'' becomes testable through local constraints and can be monitored via internal measures. The amount of contextual information carried by the unit, as well as the temporal persistence of this information, are locally tuned by distinct constraints.

This proposal addresses a structural tension: in sequential generation, causality is predominantly organized in the symbolic space and, even when latents exist, they often remain auxiliary, while the effective dynamics are carried by a largely deterministic decoder. In parallel, probabilistic latent models capture factors of variation and uncertainty, but that uncertainty typically remains borne by global or parametric mechanisms, while units continue to propagate scalars --- hence the pivot question: if uncertainty is intrinsic to computation, why does the compute unit not carry it explicitly? We therefore draw two axes: (i) the composition of probabilistic constraints, which must be made stable, interpretable, and controllable; and (ii) granularity: if inference is a negotiation of distributions under constraints, should the primitive unit remain deterministic, or become distributional? The genesis starts from multi-horizon hierarchical VAEs re-read as a composition of constraints, pushed down to the atom of probabilistic computation, motivating the VAE brick. We analyze \emph{collapse} modes and the conditions for a ``living neuron'', then extend the contribution over time via autoregressive priors over the latent, per unit.
\end{abstract}

\section{Introduction --- Where does internal computation really live?}

Dominant architectures, especially in sequential generation, organize causality and computational dynamics in the symbolic space: attention, implicit recursion, token-by-token generation. In this regime, latent variables --- when they exist --- are often auxiliary and frequently independent and identically distributed (i.i.d.), while effective dynamics are carried by a mostly deterministic decoder.

In parallel, probabilistic learning provides a natural framework to represent factors of variation and uncertainty: latent-variable models (VAEs and extensions), Bayesian models, stochastic processes. These approaches offer principled tools to quantify uncertainty and structure latent representations, but they typically locate uncertainty either at the level of a global mechanism (a latent state) or at the level of parameters (Bayesian weights).

However, a structural point remains: most networks remain systems where scalars are propagated at the unit level, while uncertainty is carried ``above'' (global latents) or ``in the weights''. This separation limits what is testable and controllable at the level of the compute brick: one may calibrate a global output, but rarely observes --- and even more rarely controls --- local probabilistic activity. This motivates the idea of a neuron whose internal state is a distribution rather than a scalar.

This highlights a simple tension: if uncertainty is intrinsic to computation, why does the compute unit not carry it explicitly? We then ask two related questions. First, how can we combine probabilistic constraints (variational experts) in a stable, interpretable, and controllable manner? Second, if inference is a negotiation of distributions under constraints, should the primitive unit remain a deterministic neuron, or become a distributional neuron?

Our hypothesis is that internal computation can be read as a contraction of a space of possibilities under local constraints: inference corresponds to the progressive reduction of a set of compatible distributions, until agreement with a prior and with observations. In this reading, variational regularizers act as constraints (or, equivalently, energy terms) that make this contraction definable and testable at the unit level.

The genesis of this work is rooted in multi-horizon hierarchical VAEs, where a hierarchy of latents can be read as a composition of probabilistic constraints. We atomize this perspective down to the elementary unit: by pushing this reading to its limit, we obtain the question of the atom of probabilistic computation, which we answer with a 1D VAE brick.

We introduce \emph{EVE} (\emph{Elemental Variational Expanse}), a variational distributional neuron: a unit formulated as a 1D VAE brick carrying an internal latent, a prior/posterior, and a local ELBO, made observable and steerable via instrumented local constraints (e.g., effective KL, a band on $\mu^2$, out-of-band fractions, drift/\emph{collapse} indicators). Beyond stabilization and anti-collapse, the band acts as a local coupling knob to context: by controlling the admissible energy or amplitude of the latent, it limits (or allows) the amount of contextual information effectively carried by the unit. This tuning is local and heterogeneous (potentially different band per neuron) and combines with an autoregressive (AR) dynamic whose strength or step can also vary per unit, defining local time scales (fast neurons vs slow neurons). To isolate these effects at the unit level, we fix the latent dimension to 1 in order to preserve an atomic variational unit and attribute the observed effects to the local variational mechanism (amortized posterior and KL regularization), rather than to a capacity gain, with minimal overhead. The paper validates this primitive and its diagnostics rather than proposing a full hierarchical architecture.

We also discuss where to place dynamics in a probabilistic model, distinguishing three levels: (i) per-unit dynamics (micro), where each unit carries its own latent trajectory (e.g., an autoregressive dynamic on a scalar), interpretable as inertia or local memory; (ii) global dynamics (macro), where dynamics live in a global latent state of \emph{state-space} type; (iii) hybrid, combining local trajectories and explicit collective coupling. In this paper, we experimentally evaluate the autoregressive extension at the neuron level, while richer extensions (e.g., Gaussian processes) are left outside the experimental scope.

\begin{figure}[htbp] 
\centering 
\includegraphics[width=\linewidth]{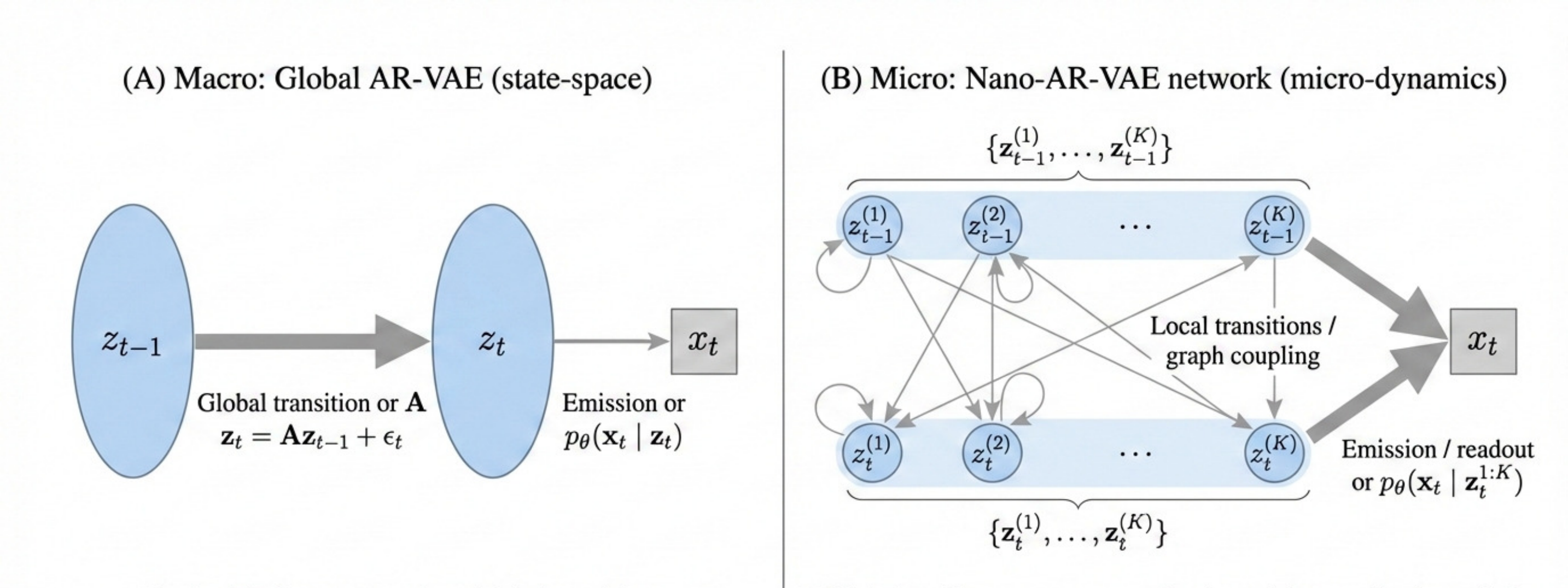} 
\caption{\textbf{Micro vs macro dynamics.} (A) Global AR-VAE: a vector latent state $z_t\in\mathbb{R}^K$ follows a state-space-like dynamics. (B) AR-VAE neuron network: local latent states $z_t^{(i)}$ evolve via local transitions (sparse graph) and are aggregated to produce the output.} \label{fig:micro_vs_macro} 
\end{figure}
\FloatBarrier

\paragraph{Contributions.}
\begin{enumerate}
    \item We define \emph{EVE} as a primitive of variational computation (1D VAE): a unit explicitly carrying a prior, a posterior, and a local ELBO.
    \item We propose a composition of probabilistic constraints observable at the unit level, including controls and diagnostics that prevent or characterize \emph{collapse} (effective KL, $\mu^2$, out-of-band fractions, drift), and we introduce a per-neuron autoregressive extension of latent dynamics.
    \item We propose a unit-centered validation protocol and targeted ablations: existence proof of a ``living'' neuron (no collapse), expressivity at comparable capacity, and long-horizon stability (i.i.d.\ prior vs AR prior), linking external performance and internal computation via per-unit latent diagnostics.
\end{enumerate}

\paragraph{Outline.}
We first formalize the hierarchy $\rightarrow$ constraints reading (experts, mechanisms, energy), then introduce the corresponding atomization (\emph{EVE}). We then evaluate the autoregressive extension at the unit level and apply the unit-centered validation protocol, before discussing limitations and perspectives.

\section{Related work}
\label{sec:related}

\paragraph{What is close and what is missing.}
Our proposal touches several existing families, without coinciding with them. We position it along four axes:
uncertainty over weights and predictions (BNN, UQ) and probabilistic neurons;
VAE and variational inference as a computation mechanism;
local objectives and \emph{layerwise} learning (intermediate criteria, composability);
sequential latent dynamics (AR / \emph{state-space}) and constraint-based control (budgets, homeostasis).

\paragraph{BNNs, uncertainty quantification, probabilistic neurons.}
BNNs, Bayesian Neural Networks, place a distribution over parameters and approximate a posterior over weights: uncertainty is carried by the weights while units remain essentially scalar \citep{blundell2015weight}. 
In uncertainty quantification (UQ), two common building blocks are calibration (e.g., \emph{temperature scaling}) \citep{guo2017calibration} and out-of-distribution (OOD) detection \citep{hendrycks2017baseline}.

An older tradition of stochastic networks (e.g., \emph{wake--sleep}) \citep{hinton1995wakesleep} is closer to the idea of a probabilistic neuron but does not formalize a composable semantics of the form ``unit = mini variational inference''.
Training through stochastic units motivates gradient estimators (REINFORCE, \emph{straight-through}, relaxations) \citep{williams1992reinforce,bengio2013stochasticneurons,jang2017gumbel,mnih2016vimco}. 
These frameworks also do not introduce a standard unit-level dashboard (effective KL, energy/$\mu^2$, drift/\emph{collapse}).

\begin{table}[htbp]
\centering
\footnotesize
\caption{Conceptual positioning. EVE moves uncertainty and inference to the unit level (local latent + micro-ELBO) and introduces local tunings/diagnostics (band/energy, effective KL, drift/\emph{collapse}), with a per-neuron AR dynamics.}
\label{tab:positioning}
\begin{tabular}{p{2cm} p{2cm} p{2cm} p{2.2cm} p{2.2cm}}
\toprule
\textbf{Approach} & \textbf{Where is uncertainty?} & \textbf{Unit} & \textbf{Objective} & \textbf{Dynamics} \\
\midrule
BNN 
& on weights 
& scalar (classic) 
& VI on weights / NLL 
& optional \\
VAE (macro) 
& global latent 
& scalar (decoder) 
& global ELBO 
& sequential prior / SSM \\
Stochastic layers 
& prediction / regularization 
& scalar + noise 
& approx.\ VI 
& weak / implicit \\
\textbf{EVE (this work)} 
& local latent per unit 
& distribution (prior/post.) 
& micro-ELBO + constraints 
& per-unit AR (micro) \\
\bottomrule
\end{tabular}
\end{table}

\paragraph{VAE and variational inference.}
VAEs connect a prior and an approximate posterior via a variational bound and the reparameterization trick \citep{kingma2014vae,rezende2014stochastic}. 
Inference is central but remains carried by global latents, while units remain deterministic.

\paragraph{Local objectives and layerwise learning.}
Local criteria have shown their interest for stabilizing layerwise training \citep{hinton2006dbn,bengio2007greedy,nokland2019local,jaderberg2017dni}. 
We propose a unit-level local ELBO (\emph{EVE}), making the internal mechanism composable and observable.

\paragraph{Latent dynamics and constraints.}
Sequential latent-variable models inject stochastic dynamics over time (broadly: latent dynamics close to \emph{state-space} models), for instance via VRNN formulations \citep{chung2015vrnn} or variants inspired by deep Kalman filters \citep{krishnan2015dkf}.
Our contribution differs by the placement of dynamics: we discuss global (macro) dynamics but mainly evaluate per-unit (micro) dynamics, and we introduce local constraints that act as budgets (band/latent energy) and stability tunings (homeostasis in the sense of controlling internal behavior), so that these dynamics remain observable and steerable.

\paragraph{Positioning: how EVE differs.}
In summary, EVE differs from the above approaches on three points:
\begin{enumerate}
    \item \textbf{Where uncertainty lives:} it is carried by the unit's internal state (local latent), not only by weights (BNN) nor only by a global latent (VAE).
    \item \textbf{What the mechanism is:} a local ELBO makes inference explicit at the unit level (``unit = mini variational inference'').
    \item \textbf{What is testable/steerable:} the proposal comes with local metrics and constraints (anti-collapse, budgets/band, dynamics diagnostics) that link external performance and internal computation and enable fine-grained control (potentially heterogeneous per neuron).
\end{enumerate}

\paragraph{What we are NOT.}
EVE is related to several existing families, but differs from them in where uncertainty, inference, and dynamics are placed. In particular, it is distinct from (i) Bayesian neural network variants, where uncertainty primarily lives in the \emph{weights} and inference targets parameters rather than unit states; (ii) generic ``stochastic layers'' (noise injection, dropout-like regularization, or stochastic activations) without an explicit latent state equipped with a prior/posterior semantics; (iii) standard variational layers in which a \emph{global} ELBO is optimized while units remain deterministic computation bricks; and (iv) state-space latent models where dynamics are centralized in a \emph{global} latent vector. EVE instead moves an explicit variational inference mechanism \emph{into the compute unit} itself and makes its internal regime measurable and controllable.

\paragraph{What is unique technically.}
The technical novelty of EVE is to define the unit as an \emph{atomic variational inference brick}: in this paper, we deliberately fix the unit latent dimension to \textbf{$k=1$} to isolate the local variational mechanism (amortized posterior + KL regularization) from a capacity increase. \textbf{With $k=1$, the contribution is not ``more latent space'', but a different unit-level semantics: computation becomes an explicit prior--posterior negotiation (micro-ELBO) with observable and controllable internal regimes.} Each neuron carries a local latent $z$, an amortized posterior $q(z\mid h)$, a prior (i.i.d.\ or AR), and a \emph{local} KL term contributing to a micro-ELBO objective. This unit-level semantics enables a unit-level \emph{dashboard} (effective KL / activity proxy $\mu^2$ / out-of-band fractions / drift indicators) and unit-level \emph{controls} (budgeted latent energy via a $\mu^2$ band, optional hard projection, and homeostatic regulation) that explicitly define internal regimes rather than silently ``regularizing'' the model. Finally, EVE extends the unit over time via \emph{per-neuron} dynamics (AR micro-dynamics) and reports corresponding internal signals (e.g., \texttt{ar\_share}), making the placement of dynamics (micro vs macro) experimentally testable.

\section{Method}
\label{sec:method}

\paragraph{Structural gap.}
The dominant literature does not provide a standard paradigm where each neuron is an explicit inference machine (prior / posterior) optimizing a composable local objective.

Our method formalizes this primitive (an atomic variational unit), as well as the mechanisms that make it stable (anti-collapse), observable (per-unit dashboard), and controllable (budgets and AutoPilot), including at large width.

\subsection{Notation and objective: from experts to the unit}
We describe a scalar variational unit and its associated local objective; the controllers introduced below make this mechanism stable and observable at large width.

\subsubsection{From variational expert to neuron}
Probabilistic models composed of variational experts reveal a structure: the global model behaves like a constraint system, where each expert exerts a ``pressure'' interpretable as an energy on the latent space. In this reading, internal computation corresponds to reducing a space of possibilities until coherence with the prior and the observations.

This raises the question of granularity: if reduction under constraints is the fundamental mechanism, why not define it locally?
We make this intuition operational by defining a compute unit able to:
\begin{enumerate}
    \item carry an internal latent state $z$,
    \item update a constraint (via a KL/prior term, or other experts),
    \item optimize a local objective (composable with other units).
\end{enumerate}
This motivates the distributional neuron: one neuron = one local step of variational inference.

\subsubsection{Notation and local objective}
\paragraph{Variables (unit level).}
\begin{itemize}
    \item Input: $h$ (or $x$)
    \item Internal latent: $z$ (scalar by default, $z\in\mathbb{R}$)
    \item Emitted message/activation: $a$
    \item Aggregated task output: $\hat y$
\end{itemize}

\paragraph{Prior and amortized posterior (unit level).}
\begin{itemize}
    \item Prior: $p(z)=\mathcal{N}(0,1)$
    \item Posterior: $q(z\mid h)=\mathcal{N}(\mu(h),\sigma(h)^2)$
    \item Reparameterization: $z=\mu(h)+\sigma(h)\varepsilon$, with $\varepsilon\sim\mathcal{N}(0,1)$
\end{itemize}

\paragraph{Local objective (minimal brick).}
\begin{equation}
\label{eq:local_objective}
L \;=\; L_{\text{task}} + \beta\,\mathrm{KL}\!\left(q(z\mid h)\,\|\,p(z)\right),
\end{equation}
where, for instance, $L_{\text{task}}=\mathrm{MSE}$.

\paragraph{Mean KL for $N$ units.}
When a layer contains $N$ units (diagonal factorization), we define:
\begin{equation}
\label{eq:kl_mean}
\mathrm{KL}_{\text{mean}}=\frac{1}{N}\sum_{i=1}^{N} \mathrm{KL}_i,
\end{equation}
to keep a stable scale as $N$ increases (cf.\ Section~\ref{sec:scaling}).

\paragraph{Added missing element.}
Unlike a classic stochastic neuron (random output without an internal latent state), EVE carries an internal latent $z$ with a prior/posterior and optimizes a local ELBO.

\subsection{EVE unit and internal measures}
\subsubsection{Operational definition}
An EVE unit implements:
\begin{enumerate}
    \item \textbf{local inference}: produce $(\mu(h),\sigma(h))$ and sample $z$ by reparameterization;
    \item \textbf{message emission}: $a\sim p(a\mid z,h)$ (or direct propagation of parameters);
    \item \textbf{regularization}: $\mathrm{KL}\!\left(q(z\mid h)\,\|\,p(z)\right)$.
\end{enumerate}

\begin{figure}[htbp]
  \centering
  \includegraphics[width=\linewidth]{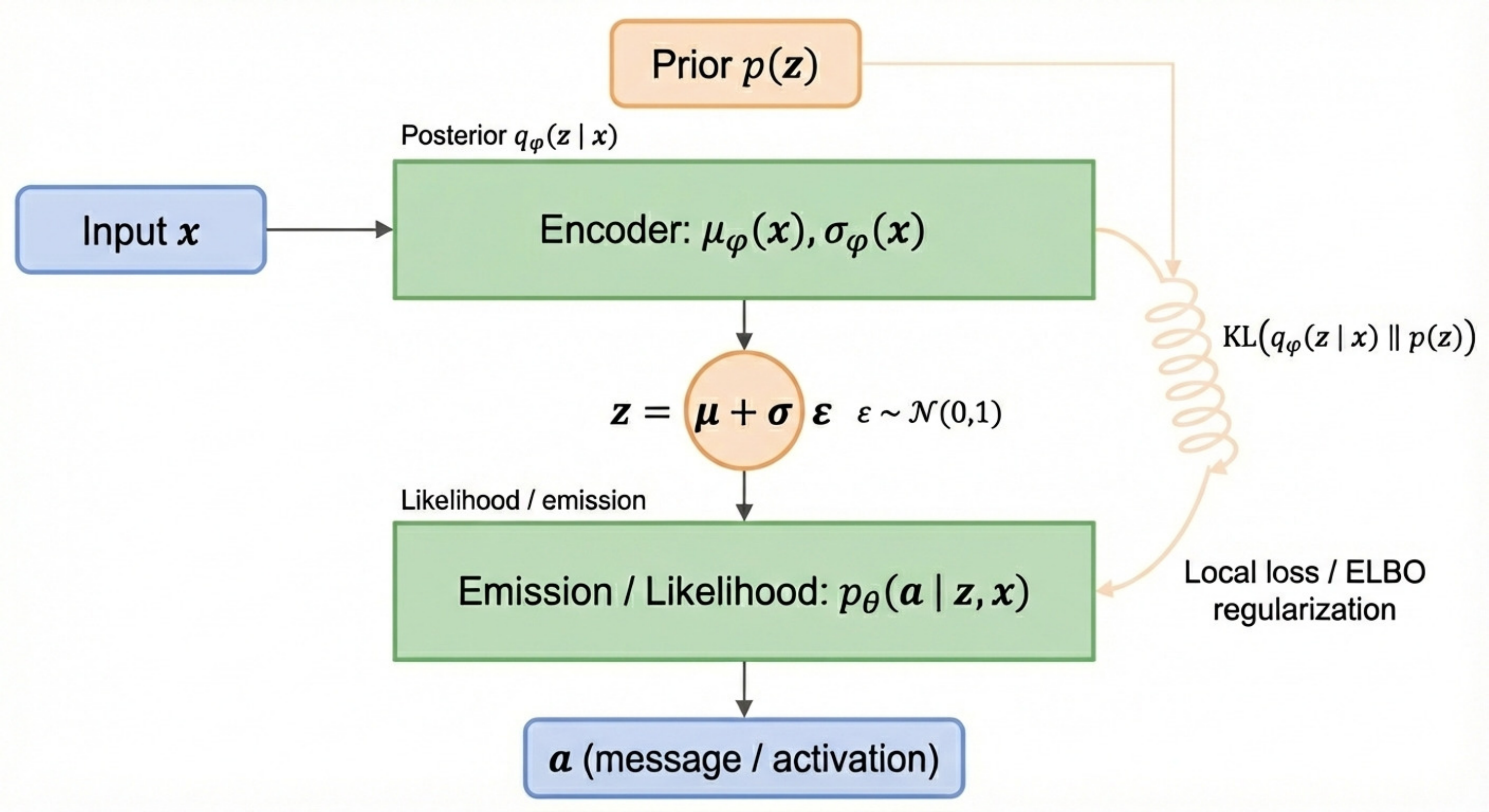}
  \caption{\textbf{Architecture of a VAE neuron (distributional neuron).}
  Each unit implements a local step of variational inference:
  $x \mapsto q_\phi(z\!\mid\!x)=\mathcal{N}(\mu_\phi(x),\sigma_\phi(x)^2)$, sampling
  $z=\mu_\phi(x)+\sigma_\phi(x)\varepsilon$, $\varepsilon\sim\mathcal{N}(0,1)$, then emission
  $p_\theta(a\!\mid\!z,x)$. The prior $p(z)$ regularizes via $\mathrm{KL}(q_\phi(z\!\mid\!x)\,\|\,p(z))$,
  defining a local ELBO (``spring'' toward the prior).}
  \label{fig:nano_vae_arch}
\end{figure}

\subsubsection{Local ELBO (generic form)}
We write a local ELBO (or local objective) in a unified form:
\begin{equation}
\label{eq:elbo_unit}
\mathrm{ELBO}_{\text{unit}}(h)
=\mathbb{E}_{q(z\mid h)}\!\left[\log p(a^{*}\mid z,h)\right]
-\beta\,\mathrm{KL}\!\left(q(z\mid h)\,\|\,p(z)\right).
\end{equation}

In the minimal instantiation used in practice, we directly use:
\begin{equation}
\label{eq:l_unit_practical}
L_{\text{unit}}=L_{\text{task}}(\hat y,y)+\beta\,\mathrm{KL}_{\text{mean}}.
\end{equation}

\subsubsection{Latent dimension k=1}
We fix the latent dimension to $k=1$ to preserve an \emph{atomic} variational unit, isolate the effect of local stochastic inference, and keep minimal overhead. Although the emitted output remains scalar, $k=1$ changes the internal structure: it is the smallest non-degenerate stochastic degree of freedom. In contrast, $k>1$ would turn the unit into a micro latent space (increased capacity); here, $k=1$ ensures that the observed effects come from the variational mechanism (local posterior + KL), not from a simple capacity increase.

\subsubsection{Chain 1: continuity with the deterministic case}
\paragraph{(A $\rightarrow$ B $\rightarrow$ C).}
(A) Define $z$ as internal latent and $a$ as emitted message.
(B) Install $q(z\mid h)$ + reparameterization + $p(a\mid z,h)$ + KL $\Rightarrow$ local objective / local ELBO.
(C) Deterministic limit: if $\sigma(h)\to 0$, then $z=\mu(h)+\sigma(h)\varepsilon \to \mu(h)$; the unit propagates a single point $\mu$ rather than a distribution.
Hence, EVE is a continuous generalization of the classic deterministic neuron (limit $\sigma\to 0$).

\begin{figure}[htbp]
  \centering
  \includegraphics[width=\linewidth]{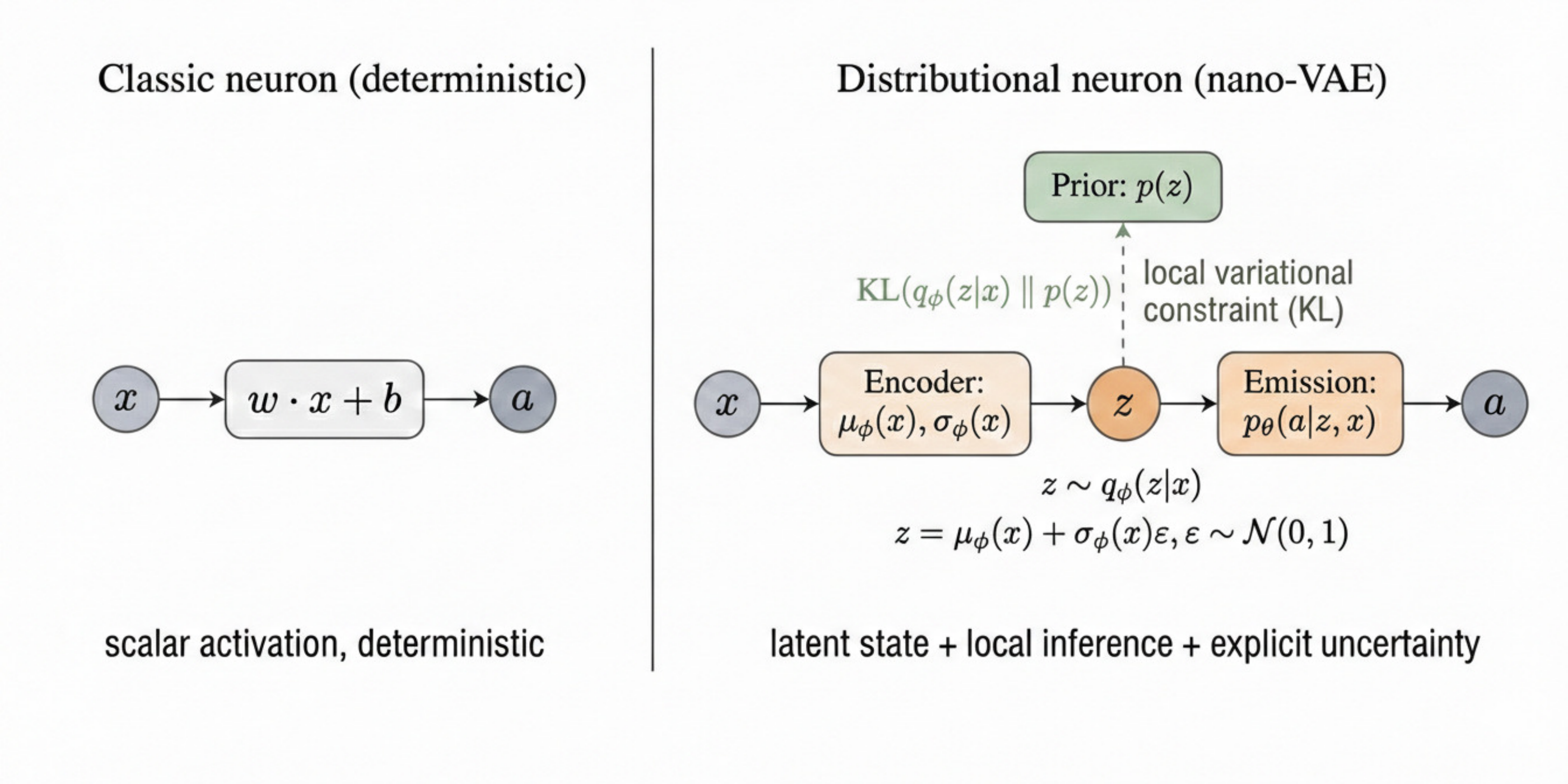}
  \caption{\textbf{Classic deterministic neuron vs distributional neuron (VAE neuron).}
  Left: deterministic scalar activation $a=w^\top x+b$.
  Right: local variational inference with internal latent $z$,
  amortized posterior $q_\phi(z\!\mid\!x)$, reparameterization, emission $p_\theta(a\!\mid\!z,x)$,
  and KL regularization. The deterministic neuron is recovered as the limit
  $\sigma_\phi(x)\to 0$.}
  \label{fig:det_vs_nano}
\end{figure}

\subsubsection{Dashboard: internal measures}
We associate each unit (or layer) with a set of observables to diagnose drift/\emph{collapse}/saturation/memory.
\begin{table}[htbp]
\centering
\scriptsize
\setlength{\tabcolsep}{2.5pt}
\caption{Internal measures (examples) and interpretation.}
\label{tab:observables}
\begin{tabular}{p{2.6cm} p{4.6cm} p{4.6cm}}
\toprule
\textbf{Measure} & \textbf{Definition / proxy} & \textbf{Interpretation} \\
\midrule
$\mathrm{KL}_{\text{mean}}$ &
$\frac{1}{N}\sum_i \mathrm{KL}_i$ &
mean information rate (structure vs fit) \\

$\mathrm{KL}_{\text{eff}}$ &
KL above a threshold, e.g., $\max(\mathrm{KL}-\tau,0)$ &
latent effectively active \\

$\mu^2_{\text{mean}}$ &
$\frac{1}{N}\sum_i \mathbb{E}[\mu_i(h)^2]$ &
energy / activity (proxy if $\sigma$ fixed) \\

low\_rate / high\_rate &
fraction of units outside budget (low/high) &
collapse / saturation / explosion \\

clamp\_low / clamp\_high &
correction amplitude (projection) &
AutoPilot intensity \\

reparam\_proxy &
$\mathbb{E}\!\left[|z-\mu|\right]$ &
consistency check (sampling) \\

AR\_share / $L_{\text{AR}}$ &
AR share / autocorrelations &
memory / inertia (vs no AR) \\
\bottomrule
\end{tabular}
\end{table}

\subsection{Constraints and controllers (AutoPilot)}
\label{sec:controllers}
This is the core of controllability: we turn the unit into an autonomous, testable system.

\subsubsection{Mini-framework: constraints = experts}
We interpret each term added to the local objective as an expert (a constraint) acting on the internal latent state. This reading unifies stability (anti-collapse), control (budgets), and diagnostics (drift, memory), because each constraint has an observable and a failure mode.

\subsubsection{Expert families and combination mechanisms}
\paragraph{Families (macro-structures).}
\begin{enumerate}
    \item By scale/horizon: \emph{coarse} constrains \emph{fine} (coarse-to-fine).
    \item By modality: a shared latent must satisfy multiple observations (PoE if simultaneous; gating if alternatives).
    \item By factors: structured factorization (spatial / temporal / causal / semantic), compatible with \emph{message passing}.
\end{enumerate}

\paragraph{Combination mechanisms.}
\begin{enumerate}
    \item Funnel (coarse-to-fine): progressive contraction.
    \item Product-of-Experts (PoE): direct intersection of constraints.
    \item Message passing: iterative dynamics toward a fixed point.
    \item Gating / mixture: mode selection/weighting.
\end{enumerate}

\subsubsection{Unified view: constraint energy}
\begin{equation}
\label{eq:energy_master}
E(z;h)=\sum_i \lambda_i\,E_i(z;h) + \beta\,E_{\text{prior}}(z),
\end{equation}
with $E_i(z;h)=-\log p_i(z\mid h)$ and $E_{\text{prior}}(z)=-\log p(z)$.
PoE corresponds to adding log-densities (intersection), the funnel to a coarse-to-fine contraction, message passing to a fixed-point dynamics, and gating to mode selection.

\subsubsection{Toward diagnostics}
Each constraint is associated with (i) an observable ($\mathrm{KL}_{\text{eff}}$, $\mu^2$, budget violations, AR share, autocorrelations) and (ii) a failure mode (collapse, saturation, drift, no-AR / AR-dominant).

\subsubsection{A budget on \texorpdfstring{$\mu^2$}{mu2}: a local informational constraint}
\label{sec:band}
\paragraph{Definition.}
For a unit $j$:
\begin{equation}
\label{eq:mu2_def}
\mu_j^2=\mathbb{E}_h\!\left[\mu_j(h)^2\right].
\end{equation}

\paragraph{Chain 2: analytic KL $\rightarrow$ $\mu^2$ energy $\rightarrow$ control.}
(A)
\[
\mathrm{KL}\!\left(\mathcal{N}(\mu,\sigma^2)\,\|\,\mathcal{N}(0,1)\right)
=
\tfrac12\left(\mu^2+\sigma^2-\log\sigma^2-1\right).
\]
(B) If $\sigma^2$ is fixed (or nearly fixed), $\mathrm{KL}\approx \tfrac12\mu^2+\text{const}$.
(C) $\mu^2$ is a direct, controllable proxy of the informational budget / latent activity.
Therefore, constraining $\mu^2$ amounts to defining a local budget per unit.

\paragraph{Heterogeneous budgets.}
We fix $[\mu^2_{\min},\mu^2_{\max}]$, then for each neuron $j$ we sample at initialization a sub-band $[\ell_j,u_j]$, with a shared central overlap (to avoid fully closing the channel for some neurons).

\subsubsection{Per-unit multiplicative projection: hard control}
After a gradient step, $\mu_j^2$ may leave $[\ell_j,u_j]$. We apply a multiplicative projection:
\begin{equation}
\label{eq:projection_mu}
(a_j,b_j)\leftarrow s_j(a_j,b_j)
\quad \Rightarrow \quad
\mu_j(h)\leftarrow s_j\mu_j(h)
\quad \Rightarrow \quad
\mu_j^2\leftarrow s_j^2\mu_j^2.
\end{equation}

\paragraph{Choice of $s_j$ (exact retraction).}
\[
s_j=
\begin{cases}
\sqrt{\ell_j/\mu_j^2} & \text{if }\mu_j^2<\ell_j,\\
\sqrt{u_j/\mu_j^2} & \text{if }\mu_j^2>u_j,\\
1 & \text{otherwise.}
\end{cases}
\]
Hence, $\mu_j^2$ is brought exactly back to the lower or upper bound when needed.

\subsubsection{Global control via \texorpdfstring{$\beta$}{beta} (optional)}
$\beta$ controls the fit-vs-structure trade-off; a $\beta$ warmup can be used if needed.

\subsubsection{AR dynamics: a local time scale}
Two levels depending on complexity:
\begin{enumerate}
    \item \textbf{Explicit AR(1) prior per unit}:
    \begin{equation}
    \label{eq:ar1_prior}
    p(z_{j,t}\mid z_{j,t-1})=\mathcal{N}(\phi_j z_{j,t-1},\sigma_{\text{AR}}^2),
    \qquad \phi_j=\exp(-\Delta t/\tau_j).
    \end{equation}
    \item \textbf{Causal AR penalty (implementation)}: add a loss $L_{\text{AR}}$ that penalizes $(\mu_t-\phi\,\mu_{t-k})$ over a stride $k$.
\end{enumerate}

\paragraph{Chain 5: budget vs time = two axes.}
Budgets control energetic amplitude (capacity) but do not fix the speed of evolution; $\tau_j$ acts via AR to control ``slow vs fast''.

\subsection{Architecture and scaling rules}
\label{sec:scaling}
\paragraph{Minimal pipeline.}
(1) encoding / input: $h$;
(2) EVE layer: $N$ units, each producing $(\mu_i(h),\sigma_i(h),z_i)$;
(3) aggregation/readout: $\hat y$ computed from statistics (often $\mu$).

\paragraph{Normalized readout (large-width stability).}
\begin{equation}
\label{eq:readout_norm}
\hat y=\frac{w^\top \mu(h)}{\sqrt{N}}+b_0.
\end{equation}

\subsection{Implementation details: optimization, logs, selection}
\paragraph{Reported global loss.}
In practice, the global loss is:
\begin{equation}
\label{eq:global_loss}
L = L_{\text{task}}(\hat y,y) + \beta\,\mathrm{KL}_{\text{mean}} + \alpha_{\text{AR}} L_{\text{AR}},
\end{equation}
with projection $\Pi_{[\ell,u]}$ applied after each update when hard control is enabled.

\subsubsection{Optimization and safeguards}
\begin{itemize}
    \item Optimizer: Adam/AdamW; scheduler if needed.
    \item Mini-batch: reduces variance, stabilizes training at large width.
    \item Monte-Carlo: typically 1 sample; sanity-check via $\mathbb{E}\!\left[|z-\mu|\right]$.
    \item Safeguards: (i) \texttt{isfinite} on loss and metrics; (ii) stats on $\mu$ (mean / max / norm); (iii) gradient clipping.
    \item Seeds: multi-seed to test robustness.
\end{itemize}

\subsubsection{Instrumented selection (LAB / score / Pareto)}
\paragraph{Principle.}
Predicting well $\neq$ obtaining the desired internal computation: MSE alone does not indicate whether the latent is used as intended. One must measure/constrain $\mu^2$, $\mathrm{KL}_{\text{eff}}$, violations, etc. We run a sweep (LAB), multi-seed, then sort by score or Pareto.

\paragraph{LAB procedure (summary).}
\begin{enumerate}
\item Sweep over: $\beta$; lr; batch size; iters; init; $\sigma$; budgets;
projection (ON / OFF); AR (ON / OFF).
    \item Multi-seed (e.g., $\{0,100,200\}$).
    \item Score (example): $\mathrm{MSE}+\lambda\cdot\max(0,\mu^2_{\text{target}}-\mu^2)^2$.
    \item Option: Pareto front (MSE vs $|\mu^2-\mu^2_{\text{target}}|$).
    \item Drop: keep top-$K$ / best percentiles.
\end{enumerate}

\section{Experiments}
\label{sec:experiments}

\paragraph{Scope choice (proof of concept).}
We deliberately adopt a single protocol (in=336, $H=96$) to provide a controlled and reproducible evaluation centered on the proposed primitive (1D VAE unit) and its controllers, rather than on dataset-specific optimization.
In this setting, we seek to \emph{isolate} and \emph{attribute} the observed effects at the unit level: existence of \textbf{observable} internal regimes (KL, $\mu^2$, \textit{out}, etc.) and \textbf{steerable} regimes (homeo/projection/AR).
An exhaustive comparison to SOTA architectures (e.g., specialized Transformers) is not the goal of this work and is a natural perspective.

\paragraph{Validation claim.}
This work does not aim at a monotone MSE improvement over an optimized deterministic baseline.
Our main claim is \emph{structural}: (i) it is possible to define a compute unit where inference and uncertainty are carried locally (prior/posterior + micro-ELBO), (ii) this internal computation becomes \emph{observable}, and (iii) it is \emph{steerable} via local constraints defining \emph{regimes} (stability/robustness vs accuracy).

In particular, we validate (a) the existence of an \emph{observable and steerable} latent regime at the unit level
(KL, $\mu^2$, \textit{out}, \texttt{band}, \texttt{ar\_share}),
and (b) a \emph{measurable trade-off} between accuracy and robustness/stability,
which a deterministic baseline cannot measure nor tune by construction.
We therefore report not only mean errors, but also their \emph{inter-seed variance} (standard deviation) and indicators of \emph{internal stability} (\textit{out}).

The experiments thus validate the existence and exploitability of these regimes and document when they become decisive (notably Weather and Exchange).

\subsection{Experimental setup}
\label{sec:exp_setup}

\paragraph{Tasks and datasets.}
We evaluate EVE on long-horizon multivariate forecasting tasks from the \textbf{LongHorizon} benchmark,
with an input window $\text{in}=336$ and a prediction horizon $H=96$.
\textbf{ECL} groups electricity consumption series with strong seasonalities and cross-correlations.
\textbf{TrafficL} contains road traffic measurements with pronounced daily/weekly patterns and regime changes.
\textbf{Weather} gathers multivariate meteorological variables, with smoother but non-stationary dynamics.
\textbf{Exchange} groups noisier exchange-rate series, serving as a demanding case study for internal computation stability.

\paragraph{Models and variants.}
At the macro level, the model is a network of EVE units (1D nVAE), which form the elementary bricks of computation.
Each unit is formulated as a mini variational inference instrumented by internal diagnostics (KL, $\mu^2$, out-of-band fraction, etc.)
to make the latent regime \emph{observable}.

We compare three internal control regimes:
(i) \textbf{homeo}: neuron-level homeostasis enabled (band regulator), without hard projection;
(ii) \textbf{projON}: hard projection enabled, without homeostasis;
(iii) \textbf{projOFF}: no internal control (neither projection nor homeostasis).
In addition, we consider a \textbf{deterministic baseline} (dedicated ablation) with identical architecture and fixed parameter budget
but \emph{without latent variables} nor KL regularization (latent internal metrics are thus undefined by construction).

\paragraph{Metrics.}
\textbf{External performance}: MSE and MAE on the test set.
\textbf{Internal diagnostics (specific to EVE)}: KL, $\mu^2_{\text{mean}}$ (mean latent energy),
$\textit{out}=\mathrm{frac}_{\mathrm{low}}+\mathrm{frac}_{\mathrm{high}}$ (fraction of units outside the stability band),
and, when logged, additional internal terms (\texttt{band}/\texttt{ar}) describing the constraint state and latent dynamics.

\paragraph{Protocol.}
All runs use the same \texttt{in}/$H$ configuration, the same training budget, and the same early-stopping scheme.
Checkpoints are selected \textbf{on validation} using a selection score (\texttt{best\_score}, mode \texttt{select\_by=score composite(MSE+internal constraints)})
and all metrics are reported \textbf{on test} at the selected checkpoint.
Unless stated otherwise, we report means $\pm$ standard deviations over $n=5$ random initializations (0--4), and all EVE results in Tables~\ref{tab:main}--\ref{tab:lh2_obs} use \textbf{AR ON}
(i.e., $\alpha_{\mathrm{AR}}>0$ in Eq.~\eqref{eq:global_loss}); Table~\ref{tab:ablation_ar} provides the dedicated AR ON/OFF ablation.

\subsection{Main results: long-horizon forecasting, internal regimes}
\label{sec:main_results}

\begin{figure}[htbp]
  \centering
  \includegraphics[width=\linewidth]{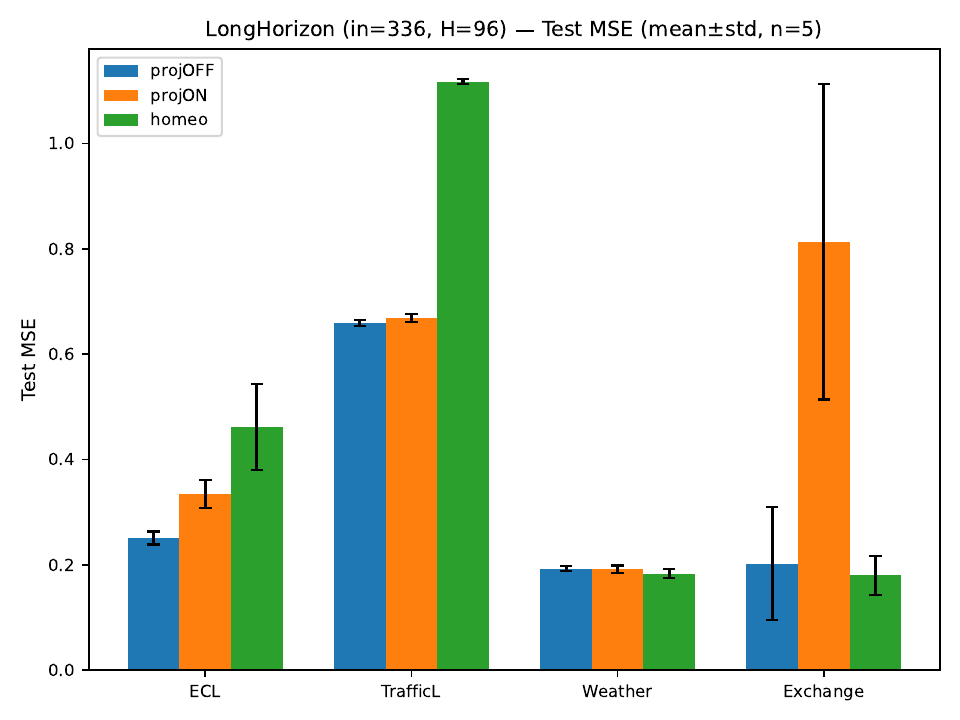}
  \caption{\textbf{LongHorizon (in=336, H=96): performance and control regimes.}
  Test MSE (mean $\pm$ std, $n=5$) for three EVE variants:
  \textbf{homeo} (homeostasis enabled), \textbf{projON} (hard projection) and \textbf{projOFF} (no internal control).
  Internal controls define \emph{regimes} and are not meant as a systematic MSE booster.}
  \label{fig:lh2_mse}
\end{figure}

\begin{table}[htbp]
{\fontsize{9}{10.8}\selectfont
\centering
\setlength{\tabcolsep}{2.6pt}
\renewcommand{\arraystretch}{0.93}
\begin{tabular}{l l r r r r}
\toprule
\shortstack{Dataset} & Variant & MSE$\downarrow$ & MAE$\downarrow$ & $\mu^2$ & \textit{out}$\downarrow$ \\
\midrule
\multirow{3}{*}{ECL ($n=5$)} & projOFF & \textbf{0.251$\!\pm\!$0.012} & \textbf{0.363$\!\pm\!$0.012} & 0.085$\!\pm\!$0.003 & \textbf{0.534$\!\pm\!$0.030} \\
 & projON & 0.335$\!\pm\!$0.026 & 0.436$\!\pm\!$0.021 & 0.049$\!\pm\!$0.002 & 0.630$\!\pm\!$0.042 \\
 & homeo & 0.461$\!\pm\!$0.082 & 0.528$\!\pm\!$0.063 & 0.133$\!\pm\!$0.088 & 0.601$\!\pm\!$0.065 \\
\midrule
\multirow{3}{*}{TrafficL ($n=5$)} & projOFF & \textbf{0.659$\!\pm\!$0.006} & \textbf{0.518$\!\pm\!$0.002} & 0.068$\!\pm\!$0.002 & \textbf{0.379$\!\pm\!$0.009} \\
 & projON & 0.669$\!\pm\!$0.008 & 0.523$\!\pm\!$0.004 & 0.064$\!\pm\!$0.001 & 0.379$\!\pm\!$0.014 \\
 & homeo & 1.117$\!\pm\!$0.005 & 0.740$\!\pm\!$0.002 & 0.082$\!\pm\!$0.002 & 0.530$\!\pm\!$0.014 \\
\midrule
\multirow{3}{*}{Weather ($n=5$)} & projOFF & 0.193$\!\pm\!$0.005 & 0.259$\!\pm\!$0.004 & 0.095$\!\pm\!$0.013 & 0.772$\!\pm\!$0.012 \\
 & projON & 0.191$\!\pm\!$0.008 & 0.259$\!\pm\!$0.003 & 0.057$\!\pm\!$0.007 & 0.783$\!\pm\!$0.012 \\
 & homeo & \textbf{0.183$\!\pm\!$0.009} & \textbf{0.247$\!\pm\!$0.005} & 0.079$\!\pm\!$0.014 & \textbf{0.573$\!\pm\!$0.022} \\
\midrule
\multirow{3}{*}{Exchange ($n=5$)} & projOFF & 0.202$\!\pm\!$0.107 & 0.331$\!\pm\!$0.095 & 0.146$\!\pm\!$0.044 & 0.594$\!\pm\!$0.056 \\
 & projON & 0.813$\!\pm\!$0.299 & 0.695$\!\pm\!$0.137 & 0.103$\!\pm\!$0.003 & 0.565$\!\pm\!$0.018 \\
 & homeo & \textbf{0.180$\!\pm\!$0.037} & \textbf{0.323$\!\pm\!$0.037} & 0.150$\!\pm\!$0.011 & \textbf{0.403$\!\pm\!$0.036} \\
\bottomrule
\end{tabular}
\caption{\textbf{LongHorizon (in=336, H=96), multi-seeds.}
\textbf{Test} metrics at the checkpoint selected \textbf{on validation} (\texttt{select\_by=score composite (MSE + internal constraints)}).
We report mean $\pm$ std over $n=5$ random initializations (0--4).
In addition to MSE/MAE, we report $\mu^2$ (mean latent energy) and \textit{out} (fraction of units outside the band,
\textit{out}=$\mathrm{frac}_{low}+\mathrm{frac}_{high}$).}
\label{tab:main}
}
\end{table}

\paragraph{Result interpretation.}
Internal controls are not intended as a universal MSE improvement: they define distinct \emph{internal regimes} (stable vs unstable, more/less constrained latents)
that MSE alone does not reveal.
Two observations are particularly robust under this protocol:
\textbf{(i) Weather:} \textbf{homeo} achieves the best mean MSE and superior internal stability (marked decrease in \textit{out}: $\approx 0.77 \rightarrow 0.57$),
and it is also the best variant in MSE \emph{for 5/5 random initializations}.
\textbf{(ii) Exchange:} \textbf{homeo} is best on average and strongly reduces inter-initialization variance (MSE std $\approx 2.9\times$ lower than \textbf{projOFF}),
while \textit{out} is lower \emph{for 5/5 random initializations}.
Conversely, on \textbf{ECL} and \textbf{TrafficL}, an overly constraining control can degrade accuracy:
\textbf{projOFF} is best in MSE (ECL: 5/5 initializations; TrafficL: 4/5 initializations), confirming that controllers should be viewed as a \emph{regime choice}.

\paragraph{Robustness and stability (beyond the mean).}
Two quantitative signals complement the mean MSE.
\textbf{(i) Inter-seed robustness:} on \textbf{Exchange}, \textbf{homeo} strongly reduces variance
(MSE $0.180\pm0.037$ vs $0.202\pm0.107$ for \textbf{projOFF}, i.e., $\approx 2.9\times$ smaller std),
suggesting a robustness effect of the controlled regime.
\textbf{(ii) Internal stability:} on \textbf{Weather} and \textbf{Exchange}, \textbf{homeo} markedly decreases \textit{out}
(Weather: $0.772\rightarrow0.573$; Exchange: $0.594\rightarrow0.403$),
indicating a more stable latent regime while preserving non-degenerate information flow (high KL, Table~\ref{tab:lh2_obs}).

\subsection{Internal observables: active and steerable latents}
\label{sec:diagnostics_obs}

External performance (MSE/MAE) is insufficient to characterize EVE's \emph{internal computation regime}.
We therefore report \emph{latent observables} measured on test, at the same checkpoint as Table~\ref{tab:main}, to make the regime \emph{testable}.

\paragraph{Active latents.}
We consider a latent regime \emph{active} when it is \emph{non - degenerate}, i.e., when the information rate does not collapse to zero.
Operationally, we declare a latent \emph{active} if $\mathrm{KL}_{\text{mean}} > \tau$, where $\tau>0$ is a small threshold fixed a priori.
When available, we also use a more robust measure,
$\mathrm{KL}_{\text{eff}} = \frac{1}{N}\sum_i \max(\mathrm{KL}_i-\tau,0)$ (cf.\ Table~\ref{tab:observables}),
and a regime is said active if $\mathrm{KL}_{\text{eff}}$ is not concentrated near zero.
In that case, the model is not operating in a ``latent ignored'' mode.

\paragraph{Steerable latents.}
A regime is said \emph{steerable} if internal controllers --- \textbf{homeo}, \textbf{projON}, and \textbf{projOFF} --- shift observables coherently without numerical pathology. We focus on the out-of-band fraction
\[
\textit{out}=\mathrm{frac}_{\mathrm{low}}+\mathrm{frac}_{\mathrm{high}},
\]
and the band penalty (\textit{Band}), which reflects the intensity and/or frequency of corrections imposed on the latent.

\begin{table}[htbp]
\centering
\scriptsize
\setlength{\tabcolsep}{3pt}
\caption{\textbf{LongHorizon (in=336, H=96) --- internal test observables (multi-seeds).}
Mean $\pm$ std over $n=5$ random initializations (0--4), reported at the same checkpoint as Table~\ref{tab:main}
(selected on validation, \texttt{select\_by=score composite (MSE + internal constraints)}).
\textbf{\textit{out}} (out-of-band): $\textit{out}=\mathrm{frac}_{low}+\mathrm{frac}_{high}$ (bold when minimal within the dataset).
$\mathrm{KL}$: latent information rate; $\mu^2$: mean latent energy; \textit{Band}: band penalty.}
\label{tab:lh2_obs}
\begin{tabular}{llcccc}
\toprule
\textbf{Dataset} & \textbf{Variant} & \textbf{\textit{out}$\downarrow$} & $\mathrm{KL}$ & $\mu^2$ & \textit{Band} \\
\midrule
ECL & projOFF & \textbf{0.534 $\pm$ 0.030} & 73.7 $\pm$ 2.7 & 0.085 $\pm$ 0.003 & 0.128 $\pm$ 0.019 \\
ECL & projON  & 0.630 $\pm$ 0.042 & 44.3 $\pm$ 9.2 & 0.049 $\pm$ 0.002 & 0.156 $\pm$ 0.010 \\
ECL & homeo   & 0.601 $\pm$ 0.065 & 113.2 $\pm$ 24.2 & 0.133 $\pm$ 0.088 & 0.217 $\pm$ 0.050 \\
\midrule
TrafficL & projOFF & \textbf{0.379 $\pm$ 0.009} & 46.3 $\pm$ 1.9 & 0.068 $\pm$ 0.002 & 0.931 $\pm$ 0.135 \\
TrafficL & projON  & 0.379 $\pm$ 0.014 & 44.3 $\pm$ 1.7 & 0.064 $\pm$ 0.001 & 1.000 $\pm$ 0.216 \\
TrafficL & homeo   & 0.530 $\pm$ 0.014 & 49.3 $\pm$ 1.4 & 0.082 $\pm$ 0.002 & 2.243 $\pm$ 0.790 \\
\midrule
Weather & projOFF & 0.772 $\pm$ 0.012 & 96.1 $\pm$ 18.8 & 0.095 $\pm$ 0.013 & 0.330 $\pm$ 0.038 \\
Weather & projON  & 0.783 $\pm$ 0.012 & 70.1 $\pm$ 23.0 & 0.057 $\pm$ 0.007 & 0.376 $\pm$ 0.059 \\
Weather & homeo   & \textbf{0.573 $\pm$ 0.022} & 171.2 $\pm$ 41.3 & 0.079 $\pm$ 0.014 & 0.237 $\pm$ 0.035 \\
\midrule
Exchange & projOFF & 0.594 $\pm$ 0.056 & 237.5 $\pm$ 22.2 & 0.146 $\pm$ 0.044 & 1.164 $\pm$ 0.637 \\
Exchange & projON  & 0.565 $\pm$ 0.018 & 297.8 $\pm$ 61.1 & 0.103 $\pm$ 0.003 & 0.701 $\pm$ 0.117 \\
Exchange & homeo   & \textbf{0.403 $\pm$ 0.036} & 253.0 $\pm$ 3.8 & 0.150 $\pm$ 0.011 & 0.574 $\pm$ 0.194 \\
\bottomrule
\end{tabular}
\end{table}

\paragraph{Reading.}
Table~\ref{tab:lh2_obs} shows distinct \emph{internal regimes}. These are defined by the controllers.
(i) On \textbf{Weather} and \textbf{Exchange}, \textbf{homeo} systematically achieves the lowest \textit{out}, indicating a more stable regime (fewer band violations) while maintaining non-degenerate latent information flow (high KL).
(ii) On \textbf{TrafficL}, \textbf{homeo} exhibits a very high \textit{Band} penalty, suggesting the controller is acting almost permanently (an ``overly constrained'' regime), consistent with the MSE degradation observed in Table~\ref{tab:main}.
(iii) On \textbf{ECL}, \textbf{projOFF} minimizes \textit{out} on average, showing that internal stability is not monotone with control strength and reinforcing the ``regime choice'' reading.

\paragraph{Scope.}
These observables show (a) the existence of effectively used latents (non-degenerate KL) and (b) the ability to \emph{steer} the regime (variation of \textit{out}/\textit{Band} across controllers), properties absent by construction in a deterministic baseline.
The next section quantifies the operational value of these signals (global correlations and early AutoPilot signals).

\subsection{Internal diagnostics: from observable to steerable}
\label{sec:diagnostics}

\paragraph{Global correlations (per-dataset normalization).}
Across all runs ($4$ datasets $\times 3$ variants $\times 5$ random initializations, $N=60$),
we compare \emph{within each dataset} (z-score per dataset to avoid scale effects).
In this setting, \textit{out} is positively correlated with test MSE (Pearson $r=0.567$, $p=2.3\times 10^{-6}$),
and KL is also positively correlated (Pearson $r=0.351$, $p=6.0\times 10^{-3}$),
while $\mu^2$ is not significantly correlated ($r=0.082$, $p=5.3\times 10^{-1}$).
\emph{Reading:} \textit{out} (and, to a lesser extent, KL) is a simple quantitative proxy of the internal regime associated with run quality,
whereas $\mu^2$ is more descriptive of the level of latent activity.

\begin{figure}[htbp]
  \centering
  \includegraphics[width=\linewidth]{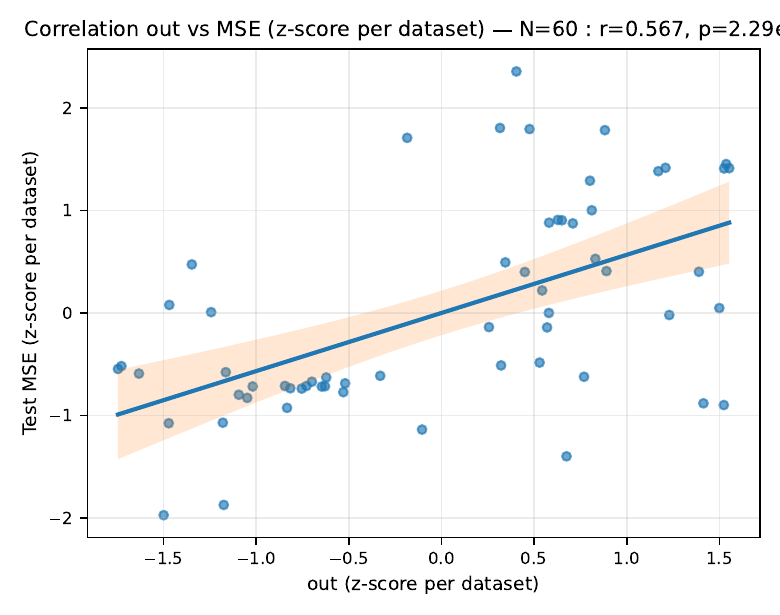}
  \caption{\textbf{Global correlation: \textit{out} vs MSE.}
  Scatter plot over $N=60$ runs (z-score per dataset), showing the positive correlation between the out-of-band fraction (\textit{out})
  and test MSE.}
  \label{fig:corr_out_mse}
\end{figure}

\paragraph{AutoPilot: early signals and selection/pruning (sweep).}
To assess the operational value of diagnostics, we perform a successive-halving sweep on \textbf{ECL} with the \textbf{homeo} variant
(24 configurations, stop at 3 epochs then pruning).
As early as 3 epochs (24 trials), internal metrics span a wide range (e.g., KL $\approx 22$ to $877$, \textit{out} $\approx 0.46$ to $0.83$, \texttt{ar\_share} $\approx 0.05$ to $0.35$),
and they are already informative:
\textit{out} is positively correlated with the best validation MSE (Pearson $r=0.582$, $p=2.8\times 10^{-3}$),
while \texttt{ar\_share} is negatively correlated ($r=-0.606$, $p=1.7\times 10^{-3}$).
\emph{Scope:} these correlations are not a universal guarantee of final performance,
but motivate using \textit{out} and \texttt{ar\_share} as early signals to filter pathological regimes
and accelerate hyperparameter search via pruning.

\begin{figure}[htbp]
  \centering
  \includegraphics[width=\linewidth]{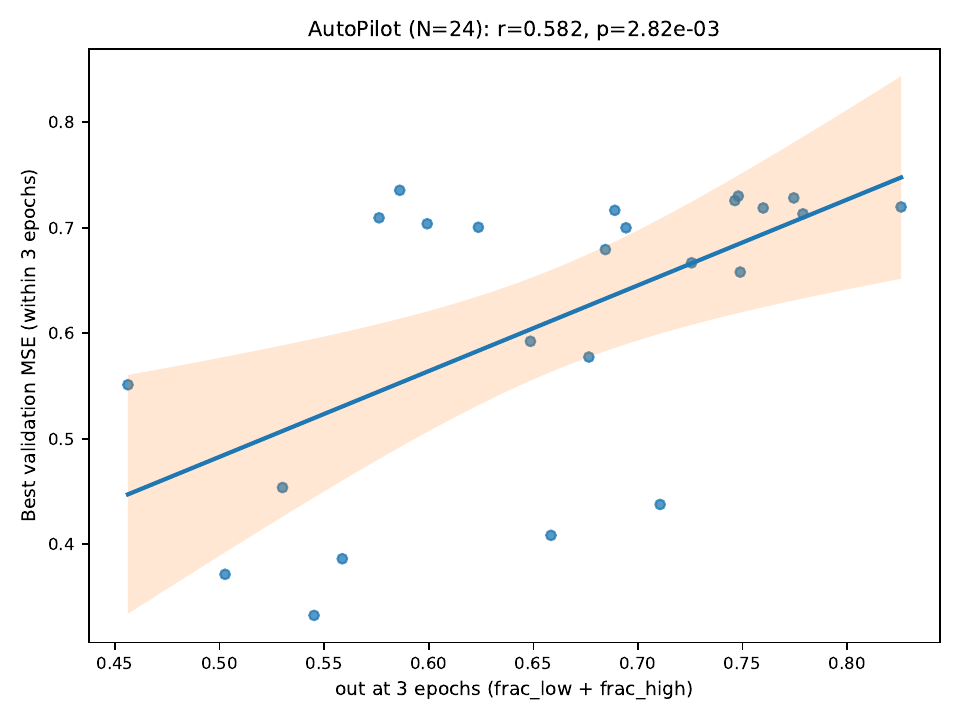}
  \caption{\textbf{AutoPilot (ECL, 24 trials, 3 epochs): early \textit{out} signal.}
  Correlation between the out-of-band fraction measured at 3 epochs and the best validation MSE reached within these 3 epochs.}
  \label{fig:AutoPilot_out}
\end{figure}

\subsection{AR ablation: ON/OFF (per-neuron dynamics)}
\label{sec:ablation_ar}

\paragraph{Goal.}
AR dynamics are a distinct contribution from budget control ( \textbf{homeo}/projection).
We isolate their effect by toggling \textbf{AR OFF} vs \textbf{AR ON} while keeping the protocol identical (same backbone, training budget, seeds, and validation-based checkpoint selection).
We report external metrics (MSE / MAE) and internal observables (\texttt{ar\_share}, \textit{out}).

\paragraph{AR share.}
We report \texttt{ar\_share} $=\alpha_{\mathrm{AR}}\,L_{\mathrm{AR}} / L_{\text{total}}$, i.e., the fraction of the training objective
contributed by the AR term (0 when AR is disabled).

\begin{table}[htbp]
\centering
\scriptsize
\setlength{\tabcolsep}{8pt}
\renewcommand{\arraystretch}{1.05}
\caption{\textbf{AR ON/OFF ablation (homeo, multi-seeds).}
Mean $\pm$ std over $n=5$ seeds (0--4). The checkpoint is selected on validation using the same composite score
(val MSE + internal constraint penalties).}
\label{tab:ablation_ar}
\begin{tabular}{l l r r r r}
\toprule
Dataset & Var. & MSE$\downarrow$ & MAE$\downarrow$ & ar\_share$\uparrow$ & out$\downarrow$ \\
\midrule
ECL      & AR OFF & 0.217$\!\pm\!$0.003 & 0.333$\!\pm\!$0.002 & 0.000$\!\pm\!$0.000 & 0.393$\!\pm\!$0.007 \\
ECL      & AR ON  & 0.461$\!\pm\!$0.082 & 0.528$\!\pm\!$0.063 & 0.367$\!\pm\!$0.219 & 0.601$\!\pm\!$0.065 \\
\midrule
Exchange & AR OFF & 0.190$\!\pm\!$0.049 & 0.331$\!\pm\!$0.043 & 0.000$\!\pm\!$0.000 & 0.392$\!\pm\!$0.042 \\
Exchange & AR ON  & 0.180$\!\pm\!$0.037 & 0.323$\!\pm\!$0.037 & 0.047$\!\pm\!$0.005 & 0.403$\!\pm\!$0.036 \\
\bottomrule
\end{tabular}
\end{table}

\paragraph{Reading.}
This ablation is not meant to seek a monotone MSE gain: it validates AR as a separate axis that changes the internal regime
(\texttt{ar\_share} and \textit{out}).
On \textbf{Exchange}, AR ON yields a slight mean improvement with moderate AR activity and comparable stability (\textit{out}).
On \textbf{ECL}, AR ON induces much stronger AR activity and a higher \textit{out}, while performance degrades sharply,
suggesting a non-trivial interaction between AR dynamics and constraints (homeo/band) that requires dedicated tuning
(e.g., $\phi$, $\sigma_{\mathrm{AR}}$, and the AR weight $\alpha_{\mathrm{AR}}$).
This negative result is informative: AR is not a ``free'' capacity booster, but a regime-changing mechanism.

\subsection{Structural ablation: deterministic vs variational unit}
\label{sec:ablation}

We directly compare the deterministic baseline to the EVE variational unit (\textbf{homeo}) under the same protocol; results are reported in Table~\ref{tab:ablation_det}.

\begin{table}[htbp]
\centering
\begingroup
\small
\setlength{\tabcolsep}{6pt}
\renewcommand{\arraystretch}{1.15}
\caption{\textbf{Dedicated ablation (multi-seeds).} Deterministic baseline vs 1D nVAE (EVE), mean $\pm$ std over $n=5$ random initializations (0--4).
The deterministic model has no latent variables; latent metrics (e.g., KL and \textit{out}) are therefore undefined.
This ablation isolates the effect \emph{``latent vs no latent''}. Here, EVE is fixed to the \textbf{homeo} regime (band regulation ON, projection OFF).}
\label{tab:ablation_det}

\begin{minipage}[t]{\linewidth}\centering
\textbf{ECL} ($n=5$)\\[2pt]
\begin{tabular}{l S[table-format=1.3(3)] S[table-format=1.3(3)]}
\toprule
Model & {MSE$\downarrow$} & {MAE$\downarrow$}\\
\midrule
Deterministic & \bfseries 0.139 \pm 0.001 & \bfseries 0.237 \pm 0.001\\
EVE (homeo)   & 0.466 \pm 0.079 & 0.531 \pm 0.060\\
\bottomrule
\end{tabular}\\[2pt]
\footnotesize\textit{EVE internals:} KL $=108.6 \pm 17.3$, \textit{out} $=0.625 \pm 0.029$.
\end{minipage}

\vspace{6pt}

\begin{minipage}[t]{\linewidth}\centering
\textbf{TrafficL} ($n=5$)\\[2pt]
\begin{tabular}{l S[table-format=1.3(3)] S[table-format=1.3(3)]}
\toprule
Model & {MSE$\downarrow$} & {MAE$\downarrow$}\\
\midrule
Deterministic & \bfseries 0.371 \pm 0.001 & \bfseries 0.301 \pm 0.001\\
EVE (homeo)   & 1.116 \pm 0.004 & 0.739 \pm 0.002\\
\bottomrule
\end{tabular}\\[2pt]
\footnotesize\textit{EVE internals:} KL $=49.2 \pm 1.5$, \textit{out} $=0.532 \pm 0.008$.
\end{minipage}

\vspace{6pt}

\begin{minipage}[t]{\linewidth}\centering
\textbf{Weather} ($n=5$)\\[2pt]
\begin{tabular}{l S[table-format=1.3(3)] S[table-format=1.3(3)]}
\toprule
Model & {MSE$\downarrow$} & {MAE$\downarrow$}\\
\midrule
Deterministic & \bfseries 0.177 \pm 0.012 & \bfseries 0.233 \pm 0.013\\
EVE (homeo)   & 0.189 \pm 0.011 & 0.259 \pm 0.014\\
\bottomrule
\end{tabular}\\[2pt]
\footnotesize\textit{EVE internals:} KL $=163.4 \pm 36.3$, \textit{out} $=0.561 \pm 0.015$.
\end{minipage}

\endgroup
\end{table}

\paragraph{Discussion and scope.}
This ablation clarifies the scope of the contribution: EVE does not aim to systematically beat a no-latent model,
but to make latent computation \emph{testable and tunable} at the unit level, via observables and controllers defining regimes.
In this setting, the deterministic baseline provides a strong reference in pure error and may dominate on some datasets;
however, it can neither measure nor control internal stability (e.g., \textit{out}) nor inter-seed robustness (standard deviation),
which become particularly informative on Weather.

\subsection{Qualitative analysis: performance vs internal regime}
\label{sec:qualitative}

\begin{figure}[htbp]
  \centering
  \includegraphics[width=\linewidth]{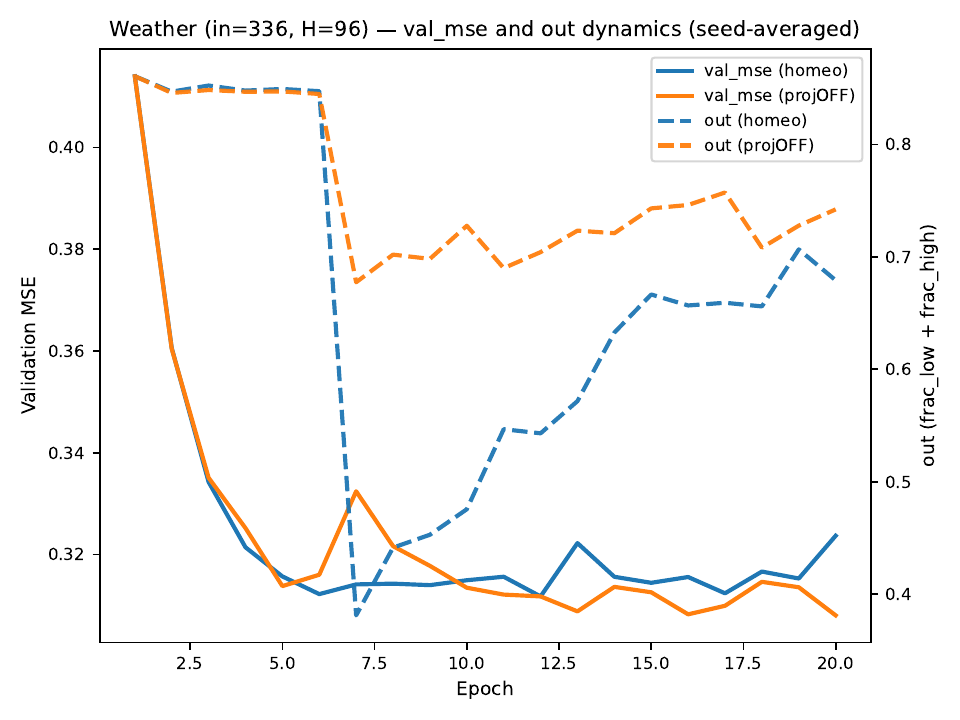}
  \caption{\textbf{Weather (in=336, H=96): training dynamics and internal regime.}
  Evolution of validation MSE (\textit{val\_mse}, left axis) and the out-of-band fraction
  (\textit{out} = \texttt{val\_frac\_too\_low} + \texttt{val\_frac\_too\_high}, right axis) for two variants:
  \textbf{homeo} (homeostasis controller) and \textbf{projOFF} (controls disabled).
  Even when MSE values remain close at some steps, internal regimes diverge:
  \textbf{homeo} typically maintains a lower \textit{out} (more stable latent regime),
  while \textbf{projOFF} operates in a regime more frequently out of band.}
  \label{fig:weather_dynamics}
\end{figure}

\paragraph{Conclusion.}
These results confirm that internal controllers define \emph{regimes} (stable vs unstable, robust vs sensitive)
that can be \emph{measured} (KL/\textit{out}/$\mu^2$) and, to some extent, \emph{exploited} (AutoPilot selection/pruning).
In this setting, EVE provides a key capability: making latent internal computation \emph{observable, quantifiable, and steerable},
and documenting when this becomes decisive (notably on Weather and Exchange, which are more sensitive to stability/robustness).

\section{Discussion}
\label{sec:discussion}

\subsection{What these experiments actually validate: a primitive and regimes}
Our experiments should not be read as a quest for a universal MSE gain.
They first validate a \emph{primitive}: a compute unit that explicitly carries internal uncertainty (prior/posterior),
optimizes a \emph{micro-ELBO}, and makes its states \emph{observable} via local metrics (KL, $\mu^2$, \textit{out}, \texttt{band}, AR signals).
The central experimental question is therefore:
\emph{can we obtain a ``living'' neuron (active, non-collapsed latent) that is controllable, at large width, with usable internal signals?}

\paragraph{Answer to the pivot question.}
Our results support that, once one wants uncertainty to be \emph{testable} and internal computation \emph{controllable}, making the unit carry a distribution (prior/posterior + micro-ELBO) is more than injected noise: it is a local semantics of inference, observable via internal metrics and controllable via constraints.
In this reading, internal computation indeed corresponds to a \emph{contraction of a space of possibilities under local constraints}, and EVE is its operational atomization.

\subsection{Internal regimes and trade-offs: accuracy vs stability and robustness}
Internal controllers (\textbf{homeo}, \textbf{projON}, \textbf{projOFF}) are not designed as MSE-optimization boosters but as mechanisms defining distinct internal \emph{regimes}.
Table~\ref{tab:main} shows that these regimes may favor either accuracy or stability / robustness, depending on the dataset.

Two observations emerge robustly under this protocol:
(i) on \textbf{Weather}, \textbf{homeo} combines the best mean MSE and a clear decrease in \textit{out} (more stable latent regime),
and it is the best MSE variant for 5/5 random initializations;
(ii) on \textbf{Exchange}, \textbf{homeo} strongly reduces inter-initialization variance and lowers \textit{out} systematically (5/5 random initializations),
suggesting a \emph{robustness regularization} effect.
Conversely, on \textbf{ECL} and \textbf{TrafficL}, an overly constraining control can degrade accuracy:
\textbf{projOFF} achieves the best MSE (ECL: 5/5; TrafficL: 4/5), confirming that control should be understood as a \emph{regime choice},
not as a monotone improvement.

\paragraph{Why \textbf{homeo} fails on TrafficL.}
On TrafficL, \textbf{homeo} exhibits a much higher band penalty (Table~\ref{tab:lh2_obs}),
indicating that the controller is activated almost permanently: the unit spends its time correcting its latent activity rather than optimizing the task.
One interpretation is that the chosen band (or its dynamics) is too restrictive for high-amplitude/high-variability patterns, reducing the effective latent capacity (a ``clamped'' regime) and degrading accuracy.
This result reinforces the idea that controllers define \emph{regimes} and must be selected (or adapted) per dataset.

\subsection{Diagnostics: why \textit{out} is an operational signal}
A practical contribution is a simple proxy of the internal regime:
\[
\textit{out}=\mathrm{frac}_{\mathrm{low}}+\mathrm{frac}_{\mathrm{high}},
\]
i.e., the fraction of units violating the band.
In our protocol (z-score per dataset, $N=60$ runs), \textit{out} is positively correlated with test MSE
(Pearson $r=0.567$, $p=2.3\times 10^{-6}$; cf.\ Section~\ref{sec:diagnostics} and Figure~\ref{fig:corr_out_mse}),
which justifies its use as a quantitative indicator of the internal regime associated with run quality.

Conversely, $\mu^2$ does not appear significantly correlated with MSE in this analysis,
which is consistent with its role: describing a level of latent activity (budget/energy), rather than a direct run-quality indicator.
In this setting, \textit{out} serves as a \emph{stability/regime} signal and $\mu^2$ as an \emph{activity-level} indicator.

\subsection{AutoPilot: selection and pruning guided by internal signals}
The correlations observed in the sweep (successive-halving) on ECL (Section~\ref{sec:diagnostics})
show that internal metrics become informative very early (as early as 3 epochs),
motivating an ``AutoPilot'' strategy: quickly filtering pathological configurations not only by external loss,
but by indicators of \emph{internal computation}.
The goal is not to guarantee final performance from a single signal,
but to accelerate hyperparameter search by early elimination of unstable regimes,
and to make selection \emph{explainable} (a run fails because it is out of band, collapsed, or dominated by an inadequate AR dynamics).

\subsection{Deterministic ablation: scope and correct interpretation}
The deterministic ablation (Table~\ref{tab:ablation_det}) shows that a no-latent model can be very competitive,
sometimes significantly better in pure error under this protocol.
This does not contradict the contribution: it clarifies the target.
EVE does not claim that ``putting a latent everywhere'' will systematically beat an optimized deterministic pipeline,
but that it is possible to move part of inference and uncertainty \emph{to the unit level} and make it
\emph{observable and steerable}.
Even when the MSE gap is small (e.g., Weather), EVE provides internal observables (KL/\textit{out}/$\mu^2$/\texttt{band}/\texttt{ar})
and control mechanisms absent by construction in the deterministic ablation.

\subsection{Per-neuron micro-dynamics (AR) vs macro dynamics: complementarity}
We distinguish (i) macro dynamics (global state-space latent), (ii) per-unit micro-dynamics (local trajectories),
and (iii) a hybrid.
Our instantiation mainly evaluates micro-dynamics via a per-unit AR(1) prior (or a causal AR penalty),
inducing local time scales (fast vs slow neurons).
This approach is compatible with richer macro architectures:
a hybrid model could combine a global state for long-range structure and micro-AR units for local memory,
all under control via budgets and diagnostics.

\subsection{Latent dimension beyond k = 1: expressivity vs controllability}
In this work, we fix $k=1$ to preserve an \emph{atomic} variational primitive and attribute observed effects
to the local inference mechanism (posterior + KL + controls), rather than to a capacity increase.
A natural extension is to study $k>1$ at the unit level, which would turn each neuron into a micro latent space:
this may increase expressivity, but raises new questions about calibration (informational budget per dimension),
stability (richer collapse modes), and instrumentation (multivariate observables, anisotropy, internal correlations).
\emph{Goal:} isolate what belongs to the \emph{local distributional semantics} (present already at $k=1$)
from what belongs to a capacity gain (which appears for $k>1$).

\subsection{Limitations and perspectives}
Several limitations are assumed in this proof of concept:
(i) the evaluation prioritizes validating a measurable/controllable latent regime rather than an exhaustive comparison to recent methods;
(ii) even with $k=1$, stochasticity introduces variance and hyperparameters (budgets, $\beta$, AR), requiring instrumented selection;
(iii) some observables (e.g., $\mu^2$) depend on implementation choices (control of $\sigma$, projection, etc.) and can be complemented.

Immediate perspectives include: adaptive budget control (learned or driven bands), hybrid micro/macro architectures,
explicit expert composition at the unit level (PoE/gating), and more systematic selection protocols (Pareto performance vs internal stability).

\paragraph{Scope of the contribution.}
In this proof of concept, EVE does not aim to replace a deterministic baseline as a universal error-minimization solution but to provide a primitive enabling one to \emph{specify} and \emph{control} probabilistic internal regimes at the unit level.
When robustness and internal stability become critical (e.g., Weather, Exchange),
these regimes may become decisive.

\section{Conclusion}
\label{sec:conclusion}

We proposed and validated a proof of concept for a \emph{variational distributional neuron}:
a unit formulated as a 1D VAE brick carrying a prior, an amortized posterior, and a local micro-ELBO.
This atomization moves part of inference and uncertainty to the unit level,
making internal computation \emph{observable} (KL, $\mu^2$, \textit{out}, \texttt{band}, AR signals) and \emph{steerable} via local controllers
(budgets/homeostasis, hard projection, per-neuron AR dynamics).

On LongHorizon (in=336, $H=96$), we show that internal controllers define \emph{regimes} that do not reduce to MSE:
on some datasets (Weather, Exchange), regulation improves internal stability and robustness,
while on others (ECL, TrafficL) a less constrained regime minimizes error.
We also establish that the out-of-band fraction \textit{out} is an operational signal:
it is correlated with test MSE in our global analysis (Section~\ref{sec:diagnostics})
and can be exploited early for AutoPilot selection/pruning in a sweep.

Finally, the deterministic ablation clarifies the scope of the contribution:
EVE does not aim to systematically beat a no-latent model,
but to demonstrate that there exists a probabilistic computation regime \emph{at the neuron level} that is stable,
instrumented, and controllable.
In summary, we (i) define a 1D VAE primitive at the unit level,
(ii) provide local anti-collapse controls/diagnostics and a per-neuron AR extension,
and (iii) experimentally validate measurable internal regimes linking internal stability and external performance.
EVE thus provides a building block for future architectures where one does not merely optimize a global error, but can also specify, diagnose, and regulate the \emph{internal dynamics} of computation.

\paragraph{Code availability.}
The code (training, configurations, analysis scripts, and figure/table generation) is available at:\\
\url{https://github.com/YvesRuffenach/EVE}.

\bibliographystyle{plainnat} 
\bibliography{references}

\end{document}